\def\paperID{0000}
\def\confName{CVPR}
\def\confYear{2024}

\def\paperTitle{Data Stream Sampling with Fuzzy Task Boundaries and Noisy Labels}

\def\authorBlock{
    Yu-Hsi Chen \\
    The University of Melbourne \\
    {\tt\small yuhsi@student.unimelb.edu.au}
}

\newif\ifreview 
\newif\ifarxiv 
\newif\ifcamera \newcommand{\cameraready}{\cameratrue}
\newif\ifrebuttal 
\cameraready 

\pdfoutput=1
\documentclass[10pt,twocolumn,letterpaper]{article}
\ifreview \usepackage[review]{cvpr} \fi
\ifarxiv \usepackage[pagenumbers]{cvpr} \fi
\ifrebuttal \usepackage[rebuttal]{cvpr} \fi
\ifcamera \usepackage{cvpr} \fi

\usepackage{graphicx}	
\usepackage{amsmath}	
\usepackage{amssymb}	
\usepackage{booktabs}
\usepackage{times}
\usepackage{microtype}
\usepackage{epsfig}
\usepackage[table,xcdraw,dvipsnames]{xcolor}
\usepackage{caption}
\usepackage{float}
\usepackage{placeins}
\usepackage{color, colortbl}
\usepackage{stfloats}
\usepackage{enumitem}
\usepackage{tabularx}
\usepackage{xstring}
\usepackage{multirow}
\usepackage{xspace}
\usepackage{url}
\usepackage{subcaption}
\usepackage{xcolor}
\usepackage{algorithm}
\usepackage{algorithmic}  
\usepackage[hang,flushmargin]{footmisc}

\ifcamera \usepackage[accsupp]{axessibility} \fi

% ...

\ifarxiv  \fi

\newcommand{\R}[1]{{%
    \textbf{%
        \ifstrequal{#1}{1}{\textcolor{red}{R#1}}{%
        \ifstrequal{#1}{2}{\textcolor{blue}{R#1}}{%
        \ifstrequal{#1}{3}{\textcolor{magenta}{R#1}}{%
        \ifstrequal{#1}{4}{\textcolor{teal}{R#1}}{%
                           \textcolor{cyan}{R#1}%
        }}}}%
    }%
}}

\newcommand{\trianglecomment}[1]{\(\triangleright\) \textit{#1}}

\DeclareMathOperator*{\argmax}{\arg\!\max}

\newcommand\independent{\protect\mathpalette{\protect\independenT}{\perp}}
\def\independenT#1#2{\mathrel{\rlap{$#1#2$}\mkern2mu{#1#2}}}  

\usepackage{xr-hyper}

\makeatletter
\newcommand*{\addFileDependency}[1]{
  \typeout{(#1)}
  \@addtofilelist{#1}
  \IfFileExists{#1}{}{\typeout{No file #1.}}
}

\makeatother

\definecolor{cvprblue}{rgb}{0.21,0.49,0.74}
\usepackage[pagebackref,breaklinks,colorlinks,citecolor=cvprblue]{hyperref}
\usepackage[capitalize]{cleveref}
\crefname{section}{Sec.}{Secs.}
\crefname{table}{Table}{Tables}
\crefname{figure}{Fig.}{Figs.}

\begin{document}
\title{\paperTitle}
\author{\authorBlock}
\maketitle

\begin{abstract}
In the realm of continual learning, the presence of noisy labels within data streams represents a notable obstacle to model reliability and fairness. We focus on the data stream scenario outlined in pertinent literature, characterized by fuzzy task boundaries and noisy labels.
To address this challenge, we introduce a novel and intuitive sampling method called Noisy Test Debiasing (NTD) to mitigate noisy labels in evolving data streams and establish a fair and robust continual learning algorithm. NTD is straightforward to implement, making it feasible across various scenarios.
Our experiments benchmark four datasets, including two synthetic noise datasets (CIFAR10 and CIFAR100) and real-world noise datasets (mini-WebVision and Food-101N). The results validate the efficacy of NTD for online continual learning in scenarios with noisy labels in data streams. Compared to the previous leading approach, NTD achieves a training speedup enhancement over two times while maintaining or surpassing accuracy levels. Moreover, NTD utilizes less than one-fifth of the GPU memory resources compared to previous leading methods.
\cameraready
The source code is available in \url{https://github.com/wish44165/ntd}.

\end{abstract}
\section{Introduction}
\label{sec:intro}

\quad Online continual learning (OCL) is the iterative process of sequentially acquiring knowledge from streaming data sources while retaining previously learned information. Unlike traditional batch learning, where models are trained on fixed datasets, continual learning systems adapt dynamically to changing data distributions and task requirements. In OCL, models must efficiently incorporate new data while mitigating catastrophic forgetting, a phenomenon where the model's performance on previous tasks deteriorates as it learns new ones. In response to this challenge, researchers have proposed various techniques, encompassing rehearsal strategies, regularization methods, and approaches focusing on parameter isolation. The significance of OCL lies in its applicability to real-world scenarios characterized by dynamic and non-stationary data environments, such as online streaming platforms, IoT devices, and autonomous systems. By enabling models to adapt incrementally to new information, OCL facilitates the development of adaptive and robust AI systems capable of continuous improvement over time.

Moreover, within the domain of OCL, scenarios are primarily classified into three main types: task incremental learning, domain incremental learning, and class incremental learning, as noted by~\cite{van2022three}. In particular, sample selection strategies~\cite{haas2016data,lughofer2017line,ramirez2017survey,prabhu2020gdumb,riemer2018learning} are fundamental components of data stream class incremental learning (CIL). The choice of sample selection strategy in CIL depends on various factors, including the data stream's characteristics, the learning task's complexity, and the learning system's computational constraints. By selecting appropriate sampling strategies, CIL algorithms can effectively manage the trade-offs between model performance, computational efficiency, and memory requirements, ultimately enabling continuous learning in dynamic and evolving environments. 

\begin{figure}[tp]
    \centering
    \includegraphics[width=\linewidth]{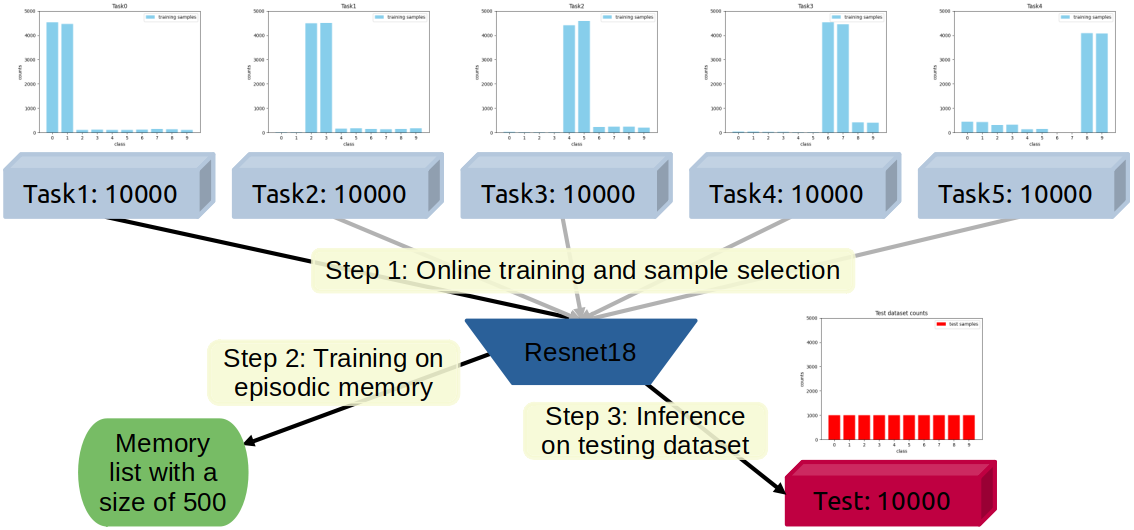}
    \caption{Illustration of the workflow for the CIFAR10 dataset with noise type Sym.-40\% and the random seed of 2.}
    \label{fig:workflow}
\end{figure}

The algorithm proposed in this paper emphasizes sample selection strategy within the context of the CIL task, considering noisy labels and fuzzy task boundaries. Fig.~\ref{fig:workflow} illustrates the comprehensive training and learning process. Our contribution can be outlined as providing the NTD algorithm, which efficiently filters out samples with correct labels and substantially reduces training time while demonstrating comparable or superior performance compared to prior approaches in the challenging and realistic continual learning scenario. The organization of this paper will unfold as follows: Sec.~\ref{sec:related} will introduce the related work, Sec.~\ref{sec:method} will detail the proposed method, Sec.~\ref{sec:experiments} will display the experimental results, and lastly, a comprehensive conclusion for the proposed method will be offered.

\section{Related Work}
\label{sec:related}

\quad In this section, we will elucidate the concepts and principles pertinent to the algorithm proposed in this paper, encompassing online continual learning, learning from noisy labels, and the imbalanced data streams in Sec.~\ref{ssec:OCL}, Sec.~\ref{ssec:bsnl}, and Sec.~\ref{ssec:ids}, respectively.

\subsection{Online Continual Learning}
\label{ssec:OCL}

\quad In the field of OCL, notable studies cover diverse research pathways and methodologies designed to address the challenges of continual learning amid dynamic environments. Some prominent strategies encompass parameter regularization~\cite{ahn2019uncertainty, jung2020continual}, function regularization~\cite{gomez2022continually}, replay techniques~\cite{shin2017continual, rolnick2019experience, kim2021continual}, context-specific components~\cite{zeng2019continual}, and template-based classification approaches~\cite{cui2021template}. In particular, the study~\cite{bang2021rainbow} underscores the significance of sample diversity within an episodic memory. Additionally, the subsequent work~\cite{bang2022online} emphasizes the importance of both sample diversity and purity within the episodic memory of continual learning models. They introduced the PuriDivER algorithm, integrating a memory management strategy that considers diversity and purity. This method incorporates semi-supervised learning into memory utilization, resulting in improved performance compared to previous approaches.

\subsection{Learning from Noisy Labels}
\label{ssec:bsnl}

\quad The challenge of handling noisy labels has spurred the exploration of innovative approaches, including developing noise-tolerant loss functions~\cite{wang2019symmetric} designed to mitigate the impact of label noise on model training. Additionally, self-supervised learning techniques~\cite{gallardo2021self} have emerged as promising solutions, leveraging intrinsic properties of data to learn representations without the need for explicit supervision. Moreover, semi-supervised approaches like PuriDivER have garnered attention in the literature, capitalizing on a combination of labeled and unlabeled data to enhance model performance and robustness. This paper will adhere to the scene settings established in PuriDivER.

\subsection{Imbalanced Data Streams}
\label{ssec:ids}

\quad Dealing with imbalanced data streams, wherein certain classes are disproportionately represented, poses a significant challenge in machine learning and data mining, notably in fields like fraud detection, anomaly detection, and network intrusion detection. Relevant strategies for managing imbalanced data streams encompass online learning~\cite{nguyen2011online,wang2014high}, incremental learning~\cite{li2020incremental,chen2011towards}, concept drift detection~\cite{korycki2021concept,wares2019data}, and active learning with sampling strategies~\cite{zhu2007active,donmez2007dual}. These methodologies signify ongoing research endeavors to overcome the hurdles of imbalanced data streams and construct resilient and adaptable machine learning models for streaming data environments.
\section{Methodology}
\label{sec:method}

\quad The proposed NTD approach uses a systematic sample selection strategy to screen out high-quality samples, enhancing accuracy during episodic memory usage. Initially, we employ the grouping method to cluster the samples. Subsequently, we utilize an augmentation method to strengthen the robustness of model predictions. Finally, data-based debiasing is applied to mitigate imbalance during the procedure. This section will delineate each component of the NTD algorithm across subsections \ref{ssec:nlg}, \ref{ssec:tta}, and \ref{ssec:dbd}, with Alg.~\ref{alg:ntd} illustrating the NTD algorithm.


\subsection{Noisy Labels Grouping}
\label{ssec:nlg}
\quad Firstly, we will cluster the samples based on the provided noisy labels while simultaneously documenting the distribution of noisy labels. This stage can considered as a preliminary phase for sample selection. Specifically, let $(x_i, \Tilde{y}_i)$ represent a sample with a noisy label pair obtained from the data stream, and let $G_c$ denote the grouping of samples associated with noisy label $c$. We can express the formulation of the noisy labels grouping as follows, given by Eq.~\ref{eq:nlg}.
\begin{equation}
\label{eq:nlg}
    G_c \cup (x_i, \Tilde{y}_i),\:\text{if}\:\Tilde{y}_i=c,\:\forall i=1,\cdots, N,
\end{equation}
where $N=|S_t|$ represents the number of samples from the data stream at task $t$.


\subsection{Test-time Augmentation}
\label{ssec:tta}

\begin{figure}[tp]
    \centering
    \includegraphics[width=\linewidth]{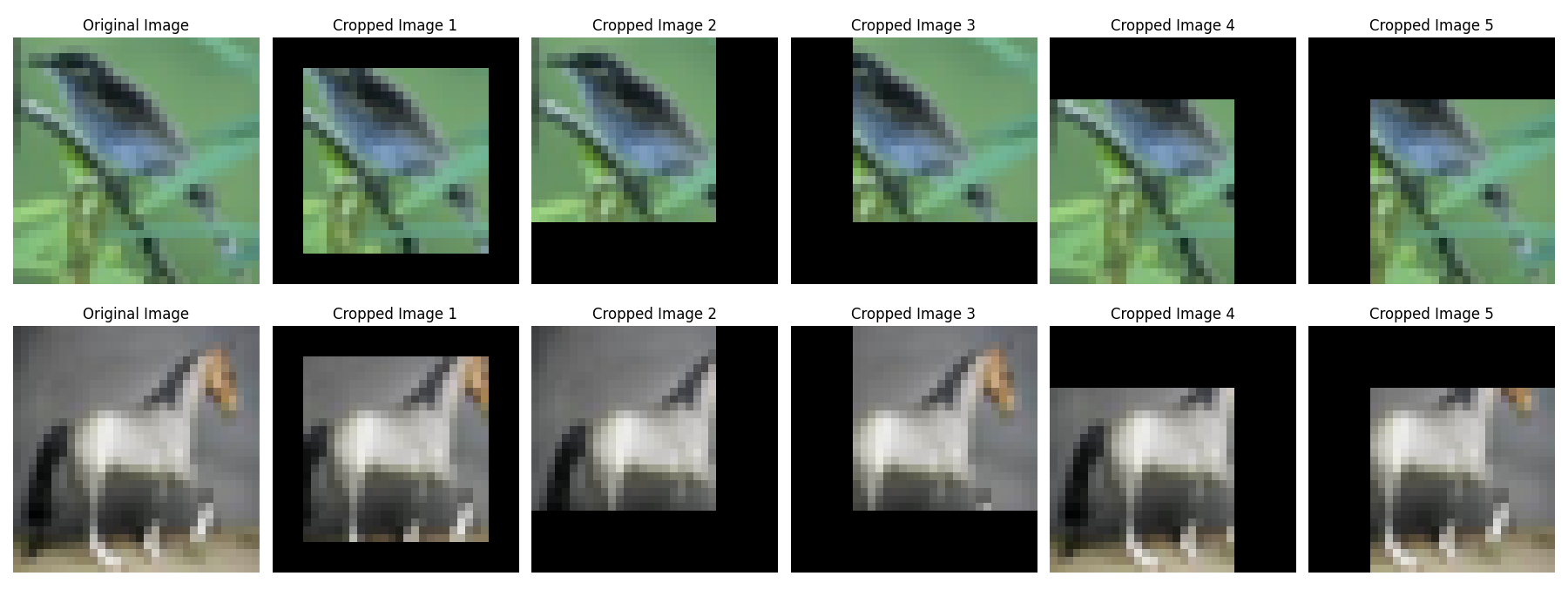}
    \caption{TTA for images with classes bird and horse.}
    \label{fig:tta}
\end{figure}

\quad When the episodic memory is fully occupied, selecting and retaining correctly annotated samples becomes critical. In this context, the measure employed is each sample's test-time augmentation (TTA) average loss value, and Fig.~\ref{fig:tta} demonstrates the visualization of TTA. Including the TTA technique in the measurement substantially improves the clean rate of the memory list. We can characterize $\Pi$ as the collection of policy-based sets encompassing various augmentation methods $\pi$ for each sample. The metric for each sample is the mean value of augmentation loss, which is computable using Eq.~\ref{eq:tta}.
\begin{equation}
\label{eq:tta}
    \Bar{\ell}_i = \frac{1}{|\Pi|}\sum\limits_{\pi \in \Pi}\ell (\pi(x_i), \Tilde{y}_i),\:\forall i=1,\cdots, N,
\end{equation}
where $\ell$ is the loss function consistent with online training.


\subsection{Data-based Debiasing}
\label{ssec:dbd} 

\quad During the construction of the episodic memory, we also employ data-based debiasing to ensure, as much as possible, an equal quantity of each category within the memory list. This alleviates data imbalance, thereby maintaining the model's performance during testing. As demonstrated in Eq.~\ref{eq:dbd}, the sample with the highest mean loss within the group with the largest number of samples is removed from the memory list during this step. This action aims to balance the number of samples across each class and subsequently alleviate bias from the training process.
\begin{equation}
\label{eq:dbd}
    G_c \setminus \argmax\limits_{(x_i, \tilde{y}_i)\in G_k} \Bar{\ell}(x_i, \tilde{y}_i),\:\text{if}\:|\mathcal{M}|>k,
\end{equation}
where $k$ is the prespecified episodic memory size. We may also consider the labeled information as an attribute that ought to have an equal probability for the chosen sample in episodic memory, as depicted in Eq.~\ref{eq:dp}, resembling a demographic parity procedure within the fairness domain~\cite{corbett2023measure}.
\begin{equation}
\label{eq:dp}
    D \independent c,
\end{equation}
where $D$ represents the distribution of decisions in the episodic memory and $c$ signifies the labeled information.

\begin{algorithm}
\footnotesize 
    \caption{Noisy Test Debiasing Sampling}
    \begin{algorithmic}[1]
    \label{alg:ntd}
    \STATE{{\bf Input:} $S_t$: data stream at task $t$, $T$: total number of tasks, $\theta_0$: initial model, $k$: size of episodic memory.}
    \STATE{{\bf Output:} $\mathcal{M}$: episodic memory.}
    \FOR{$t = 1$ to $T$}
        \FOR{mini-batch $\mathcal{B} \in S_t$}
            \STATE{$\theta \leftarrow \theta - \gamma\nabla\sum_{(x_i, \Tilde{y}_i)\in \mathcal{B}}\ell(\theta(x_i), \Tilde{y}_i)$} \hspace*{\fill}\trianglecomment{OL}
            \FOR{$(x_i, \Tilde{y}_i)\in \mathcal{B}$}
                \IF{$|\mathcal{M}| < k$}
                    \STATE{$\mathcal{M} = \mathcal{M} \cup (x_i, \Tilde{y}_i)$}
                \ELSE
                    \STATE{\texttt{Noisy Labels Grouping}} \hspace*{\fill}\trianglecomment{Sec.\ref{ssec:nlg}}
                    \STATE{\texttt{Test-time Augmentation}} \hspace*{\fill}\trianglecomment{Sec.\ref{ssec:tta}}
                    \STATE{\texttt{Data-based Debiasing}} \hspace{18.8mm}\trianglecomment{Sec.\ref{ssec:dbd}}
                \ENDIF
            \ENDFOR
        \ENDFOR
        \RETURN $\mathcal{M}$
    \ENDFOR
    \end{algorithmic}
    \label{alg:adam}
\end{algorithm}

The proposed algorithm focuses solely on constructing episodic memory without any additional processing steps during the usage of episodic memory. Therefore, our algorithm demonstrates considerably faster than the previous leading approach. Detailed experimental demonstrations will be presented in Sec.~\ref{sec:experiments}.
\section{Experiments}
\label{sec:experiments}

\quad This section will delineate the evaluation criteria, describe the benchmark dataset, outline the hyperparameters employed for the deep learning models, and present the experimental outcomes.

\subsection{Evaluation Criteria}
\label{ssec:criteria}

\quad In the continual learning experiments, we maintain alignment with the established evaluation standards prevalent in the relevant literature, which is the last test accuracy. Additionally, we record the proportion of clean data stored in the episodic memory, providing insights into the model's capacity to preserve accurate information over time. Moreover, we compare the time spent and monitor GPU memory usage throughout the training procedure.

\subsection{Dataset Descriptions}
\label{ssec:datasets}

\quad Our experiments showcase the last test accuracy and last memory clean ratio across four datasets: CIFAR10, CIFAR100, mini-WebVision, and Food-101N. Tab.~\ref{tab:dataset_hyperparameter} presents the training and validation data quantities, the number of categories, memory size constraints, the utilized models, and the corresponding hyperparameter settings during training for each dataset. Specifically, the term "symmetric" (abbreviated as "Sym.") denotes a scenario where the existence of noise in the label is independent of the true label. Conversely, "asymmetric" (abbreviated as "Asym.") denotes a condition where the noise depends on the class.

\begin{table}
\centering
\footnotesize
  \caption{Descriptions for datasets and hyperparameters.}
  \label{tab:dataset_hyperparameter}
  \begin{tabular}{lcccc}
    \toprule
    Datasets & CIFAR10 & CIFAR100 & WebVision & Food-101N \\
    \midrule
    \# train & $50000$ & $50000$ & $65944$ & $52867$ \\
    \# test & $10000$ & $10000$ & $2500$ & $4741$ \\
    \# class & $10$ & $100$ & $50$ & $101$ \\
    \# tasks & $5$ & $5$ & $10$ & $5$ \\
    Memory size & $500$ & $2000$ & $1000$ & $2000$ \\
    Models & ResNet18 & ResNet32 & ResNet34 & ResNet34 \\
    Batch size & $16$ & $16$ & $16$ & $16$ \\
    Epochs & $256$ & $256$ & $128$ & $128$ \\
    \bottomrule
  \end{tabular}
\end{table}


\subsection{Implementation Details}
\label{ssec:details}

\quad To ensure robust validation, we performed all experiments three times on a laptop featuring a CPU model 12th Gen Intel Core i7-12650H, GPU model NVIDIA GeForce RTX 4050, and 24GB of memory. Following that, we provide the average results in each cell as $\mu \pm \sigma$, where $\mu$ and $\sigma$ represent the mean and standard deviation values across three trials.

\subsection{Results}
\label{ssec:results}

\quad In the analysis, we consider the PuriDivER approach, which performs the best in the prior works, as the baseline. The experimental outcomes detailed in Tab.~\ref{tab:last_test_accuracy_cifar} and Tab.~\ref{tab:last_test_accuracy_food101n_webvision} portray the last test accuracy across four datasets. These tables demonstrate that the proposed NTD algorithm can achieve results comparable to the baseline on the CIFAR10 and CIFAR100 datasets. Meanwhile, the last test accuracy of NTD notably improves on more complex and realistic noise datasets mini-WebVision and Food-101N, with improvements reaching $1\%$ and $3.2\%$, respectively.

Moreover, we compared the clean ratio for episodic memory across different methodologies, as outlined in Tab.~\ref{tab:last_memory_clean_ratio_cifar} and Tab.~\ref{tab:last_memory_clean_rate_food101n_webvision}. The findings indicate that the utilization of NTD results in an elevated overall clean ratio within episodic memory in contrast to the baseline, especially showcasing an augmentation of $8.2\%$ on the CIFAR10 with Asym-40\%. Through the comparison of clean ratios, it is also evident that when the clean ratio is sufficiently high, it prevents the model from being misled during episodic memory usage. Lastly, based on the data presented in Tab.~\ref{tab:training_time} and Tab.~\ref{tab:gpu_memory_usage}, it can be inferred that NTD achieves a processing speed of $2.3$ times faster than the baseline for the overall training procedure while consuming less than one-fifth of the GPU memory resources during the episodic memory usage stage.

\begin{table*}
\centering
\footnotesize
  \caption{Last test accuracy evaluated on CIFAR10 and CIFAR100 datasets with noisy types Sym.-{20\%, 40\%, 60\%} and Asym.-{20\%, 40\%}.}
  \label{tab:last_test_accuracy_cifar}
  \begin{tabular}{@{}lcccccccccc@{}}
    \toprule
    \multirow{4}{*}{Methods} & \multicolumn{5}{c}{CIFAR10} & \multicolumn{5}{c}{CIFAR100}\\
    \cmidrule(lr){2-6} \cmidrule(lr){7-11}
    & \multicolumn{3}{c}{Sym.} & \multicolumn{2}{c}{Asym.} & \multicolumn{3}{c}{Sym.} & \multicolumn{2}{c}{Asym.}\\
    & 20 & 40 & 60 & 20 & 40 & 20 & 40 & 60 & 20 & 40\\ 
    \cmidrule(r){1-1} \cmidrule(lr){2-4} \cmidrule(lr){5-6} \cmidrule(lr){7-9} \cmidrule(lr){10-11}
    PuriDivER~\cite{bang2022online} & $\boldsymbol{60.6}_{\pm 1.8}$ & $57.8_{\pm 2.2}$ & $\boldsymbol{52.0}_{\pm 2.8}$ & $\boldsymbol{61.2}_{\pm 2.9}$ & $49.4_{\pm 5.7}$ & $36.3_{\pm 0.3}$ & $33.6_{\pm 0.7}$ & $\boldsymbol{28.6}_{\pm 1.7}$ & $34.3_{\pm 1.0}$ & $24.9_{\pm 1.4}$ \\
    \textbf{NTD} (ours) & $59.8_{\pm 0.6}$ & $\boldsymbol{59.7}_{\pm 1.5}$ & $50.9_{\pm 0.3}$ & $60.1_{\pm 0.3}$ & $\boldsymbol{53.7}_{\pm 3.9}$ & $\boldsymbol{38.3}_{\pm 1.0}$ & $\boldsymbol{35.2}_{\pm 1.5}$ & $28.0_{\pm 1.1}$ & $\boldsymbol{37.6}_{\pm 0.3}$ & $\boldsymbol{25.8}_{\pm 1.0}$ \\
    \bottomrule
  \end{tabular}
\end{table*}

\begin{table*}
\centering
\footnotesize
  \caption{Last memory clean ratio on CIFAR10 and CIFAR100 datasets with noisy types Sym.-{20\%, 40\%, 60\%} and Asym.-{20\%, 40\%}.}
  \label{tab:last_memory_clean_ratio_cifar}
  \begin{tabular}{@{}lcccccccccc@{}}
    \toprule
    \multirow{3}{*}{Methods} & \multicolumn{5}{c}{CIFAR10} & \multicolumn{5}{c}{CIFAR100}\\
    \cmidrule(lr){2-6} \cmidrule(lr){7-11}
    & \multicolumn{3}{c}{Sym.} & \multicolumn{2}{c}{Asym.} & \multicolumn{3}{c}{Sym.} & \multicolumn{2}{c}{Asym.}\\
    & 20 & 40 & 60 & 20 & 40 & 20 & 40 & 60 & 20 & 40\\ 
    \cmidrule(r){1-1} \cmidrule(lr){2-4} \cmidrule(lr){5-6} \cmidrule(lr){7-9} \cmidrule(lr){10-11}
    PuriDivER~\cite{bang2022online} & $98.6_{\pm 0.7}$ & $96.1_{\pm 0.6}$ & $86.6_{\pm 4.0}$ & $98.7_{\pm 0.4}$ & $79.7_{\pm 7.7}$ & $\boldsymbol{99.2}_{\pm 0.1}$ & $\boldsymbol{95.5}_{\pm 0.8}$ & $80.8_{\pm 2.3}$ & $\boldsymbol{97.0}_{\pm 0.2}$ & $\boldsymbol{72.6}_{\pm 2.1}$ \\
    \textbf{NTD} (ours) & $\boldsymbol{99.2}_{\pm 0.5}$ & $\boldsymbol{97.1}_{\pm 0.9}$ & $\boldsymbol{86.8}_{\pm 0.7}$ & $\boldsymbol{98.7}_{\pm 1.0}$ & $\boldsymbol{87.9}_{\pm 4.6}$ & $99.0_{\pm 0.3}$ & $94.8_{\pm 1.3}$ & $\boldsymbol{82.1}_{\pm 1.2}$ & $96.2_{\pm 0.3}$ & $72.1_{\pm 2.2}$ \\
    \bottomrule
  \end{tabular}
\end{table*}

\begin{table}
\centering
\footnotesize
 \caption{Last test accuracy evaluated on WebVision and Food-101N.}
 \label{tab:last_test_accuracy_food101n_webvision}
 \begin{tabular}{lcc}
   \toprule
   Methods & WebVision & Food-101N \\
   \midrule
   PuriDivER~\cite{bang2022online} & $25.1_{\pm 0.8}$ & $13.8_{\pm 0.6}$\\
   \textbf{NTD} (ours) & $\boldsymbol{26.1}_{\pm 1.6}$ & $\boldsymbol{17.0}_{\pm 0.9}$ \\
   \bottomrule
 \end{tabular}
\end{table}

\begin{table}
\centering
\footnotesize
 \caption{Last memory clean ratio on WebVision and Food-101N.}
 \label{tab:last_memory_clean_rate_food101n_webvision}
 \begin{tabular}{lcc}
   \toprule
   Methods & WebVision & Food-101N \\
   \midrule
   PuriDivER~\cite{bang2022online} & $100_{\pm 0}$ & $100_{\pm 0}$\\
   \textbf{NTD} (ours) & $\boldsymbol{100}_{\pm 0}$ & $\boldsymbol{100}_{\pm 0}$\\
   \bottomrule
 \end{tabular}
\end{table}

\begin{table}
\centering
\footnotesize
 \caption{The average training time on the CIFAR10 dataset with noisy type Sym.-40\% across three distinct random seeds for the online learning stage, the episodic memory usage stage, and the overall process (measured in hours).}
 \label{tab:training_time}
 \begin{tabular}{lccc}
   \toprule
   Methods & Online learning & Episodic memory usage & Overall\\
   \midrule
   PuriDivER~\cite{bang2022online} & $0.28$ & $3.09$ & $3.37$\\
   \textbf{NTD} (ours) & $\boldsymbol{0.19}$ & $\boldsymbol{1.25}$ & $\boldsymbol{1.44}$ \\
   \bottomrule
 \end{tabular}
\end{table}

\begin{table}
\centering
\footnotesize
 \caption{GPU memory usage for the CIFAR10 dataset with noisy type Sym.-40\% during the online learning and episodic memory usage stages (measured in MiB).}
 \label{tab:gpu_memory_usage}
 \begin{tabular}{lcc}
   \toprule
   Methods & Online learning & Episodic memory usage \\
   \midrule
   PuriDivER~\cite{bang2022online} & $828$ & $4528$\\
   \textbf{NTD} (ours) & $\boldsymbol{828}$ & $\boldsymbol{834}$ \\
   \bottomrule
 \end{tabular}
\end{table}

\section{Conclusion}
\label{sec:conclusion}

\quad The research presented in this paper centers on the data stream sampling strategy. Utilizing NTD, the method aims to augment the proportion of accurately labeled samples stored within the episodic memory, facilitating enhanced learning dynamics during the episodic memory usage phase. Empirical findings demonstrate that the last test accuracy on CIFAR datasets is comparable to the baseline approach, exhibiting pronounced enhancements across intricate and authentic noise datasets. Moreover, contrasted with the baseline, the method showcases markedly superior performance in clean ratio metrics. Furthermore, considering factors such as training time and hardware requisites, the proposed approach distinctly surpasses the baseline model.

Additionally, NTD provides multiple benefits. Its intuitive and user-friendly design facilitates effortless deployment across diverse environments, and using NTD eliminates the need for parameter adjustments that may introduce human biases. Also, its swiftness and tiniest resource requirements make it well-suited for integration into edge computing environments, aligning effectively with societal needs.
\section{Acknowledgements}
\label{sec:acknowledgements}

\quad We extend our gratitude to the authors referenced in \cite{bang2022online} for furnishing the organized code base, facilitating the reproducibility of results, and enabling performance comparison with multiple approaches.

{\small
\bibliographystyle{ieeenat_fullname}
\bibliography{11_references}
}

\ifarxiv \clearpage \appendix \section{Appendix Section}
Supplementary material goes here.
 \fi

\end{document}


\title{\paperTitle}
\author{\authorBlock}
\maketitlesupplementary

\section{Appendix Section}
Supplementary material goes here.

{\small
\bibliographystyle{ieee_fullname}
\bibliography{11_references}
}